%% file: main_arxiv.tex
\documentclass[letterpaper]{article} 
\usepackage[preprint]{aaai2027}
\usepackage[hyphens]{url}  
\usepackage{graphicx} 
\usepackage{natbib}  
\usepackage{caption} 
\usepackage{algorithm}
\usepackage{algorithmic}

\usepackage{newfloat}
\usepackage{listings}
\DeclareCaptionStyle{ruled}{labelfont=normalfont,labelsep=colon,strut=off} 
\floatstyle{ruled}
\newfloat{listing}{tb}{lst}{}
\floatname{listing}{Listing}

\usepackage{booktabs}
\usepackage{amsfonts}       
\usepackage{nicefrac}       
\usepackage{microtype}      
\usepackage{xcolor}         
\usepackage{arydshln}
\usepackage{comment}

\usepackage{mathtools}

 \usepackage{enumitem}
 \usepackage{amsmath}
 \usepackage{amssymb}
 \usepackage{graphicx}
 \usepackage{multirow}
\usepackage{adjustbox}
\usepackage{threeparttable}

\title{Foveated Probes Recover Localized Binding Information \\in Vision Foundation Models}
\author{
    Mateusz Michalkiewicz\textsuperscript{\rm 1},
    Mahsa Baktashmotlagh\textsuperscript{\rm 2},
    Guha Balakrishnan\textsuperscript{\rm 1}
}
\affiliations{
    \textsuperscript{\rm 1}Rice University, Houston, TX, USA\\
    \textsuperscript{\rm 2}The University of Queensland, Brisbane, QLD, Australia\\
}

\begin{document}

\maketitle

\input{0-abstract}

\input{1-introduction}
\input{3-method}

\input{4-tasks_and_datasets}
\input{5-experiments}

\input{6-conclusion}

\clearpage

\noindent\textbf{Acknowledgement.} Supported by the Intelligence Advanced Research Projects Activity (IARPA) via Department of
Interior/ Interior Business Center (DOI/IBC) contract number 140D0423C0076. The U.S.
Government is authorized to reproduce and distribute reprints for Governmental purposes
notwithstanding any copyright annotation thereon. Disclaimer: The views and conclusions
contained herein are those of the authors and should not be interpreted as necessarily
representing the official policies or endorsements, either expressed or implied, of IARPA,
DOI/IBC, or the U.S. Government.

\bibliography{aaai2027}

\clearpage
\newpage

\input{appendix}


\end{document}

%% file: 0-abstract.tex
\begin{abstract}
Frozen vision foundation models are commonly evaluated through a single global
image embedding, but this interface can conflate missing information with
information lost at readout time. We study this distinction by keeping a
pretrained vision encoder frozen and varying only the readout applied to its
final patch tokens. We compare standard global readouts against a
lightweight foveated readout, which attention-pools patch tokens using a
learned or question-conditioned query, and against an oracle readout with
access to the annotated target region. We evaluate these interfaces on three
localized binding problems: a controlled synthetic color--shape binding
task under clutter, a color-free crowded shape-detection variant, and a
GQA-derived natural-image task where paired questions ask for the colors
of different same-category objects in the same image.
Global readouts perform near perfectly when the synthetic target appears
alone, but collapse under clutter and counterfactual target edits, whereas
the foveated readout recovers most of the oracle-accessible signal. On the
GQA-derived task, question-independent global image vectors improve only
modestly over question-only priors, while question-conditioned foveation
substantially improves paired localized color accuracy.
A counterfactual nuisance-to-signal
ratio explains the synthetic failures: global pooling dilutes localized
label-changing evidence while exposing the probe to nuisance variation from
irrelevant objects. These results indicate that apparent spatial blindness
in frozen vision models can arise from the global embedding interface
rather than from an absence of spatial information in the frozen patch
tokens.
\end{abstract}



%% file: 1-introduction.tex
\section{Introduction}
\label{sec:introduction}

Probing is the standard tool for evaluating what vision foundation models
encode: a lightweight probe is fit on frozen features for a downstream
classification task, and the probe's accuracy is read as evidence about the
encoder. The predominant interface is a single global vector per image---a
class token, a pooled output, or a global average over patch tokens---a
natural default for contrastive vision--language encoders such as CLIP and
SigLIP \cite{radford2021learning,zhai2023sigmoid}, which are trained around
global image--text alignment.
This protocol is convenient, but it introduces a critical ambiguity: when
such a probe yields low accuracy on a spatial, relational, or binding
task, is the relevant information absent from the frozen encoder, or was
it simply discarded by the readout?

A large body of work documents these failures: vision--language models
evaluated through global image--text alignment struggle with compositional
matching, attribute--object binding, word order, and localized spatial
relations \cite{thrush2022winoground,yuksekgonul2022and,liu2023visual,
kamath2023whatsup,subramanian2022reclip}. Targeted studies directly
examine concept binding and multi-object interference
\cite{lewis2024does,campbell2024understanding}. At the same time,
dense-prediction methods demonstrate that CLIP-like patch tokens carry
useful local information despite image-level training
\cite{zhou2022extract,rao2022denseclip,wang2024sclip}, and recent analyses
suggest binding information can survive within a single modality even when
global cross-modal alignment ignores it \cite{koishigarina2025clip}.
Together, these findings suggest that a global embedding may appear
spatially blind even when the underlying patch tokens are not.

We call this possibility \emph{readout-level spatial blindness}: localized
information is present in the frozen patch tokens but becomes inaccessible
once they are pooled into a global vector, as opposed to a genuine
representation failure, in which the information is absent altogether. The
distinction is sharpest for binding: in a cluttered scene, the correct
label may depend on the color--shape conjunction of one small object, or a
question may ask for the color of one of several same-category instances.
A question-independent global readout collapses the scene before any
task-specific selection occurs, attenuating the label-relevant signal
while mixing in nuisance variation from irrelevant objects.

We separate the two cases by holding the encoder fixed and varying only
the readout; no gradients update the vision or text towers. We compare
global average pooling (GAP) and the encoder's pretrained summary
embedding against a minimal \emph{foveated} readout, which attention-pools
the patch tokens using a single query---learned in the vision-only tasks,
derived from the frozen question embedding in the vision--language
task---so that evidence is selected \emph{before} spatial collapse. We
also include a non-deployable oracle readout that pools tokens within the
annotated target region. If the oracle succeeds where the global readout
fails, the information was present in the tokens; if the learned fovea
also succeeds, a lightweight selector can recover it without fine-tuning
the encoder or any location supervision during training.

We evaluate these interfaces on three localized binding tasks: a
controlled synthetic color--shape binding task under clutter, motivated by
feature-integration theory \cite{treisman1980feature,treisman1982illusory};
a color-free crowded shape-detection variant; and a GQA-derived
natural-image task \cite{hudson2019gqa,krishna2017visual} in which paired
questions ask for the colors of different same-category objects in the same
image.

On paired-counterfactual accuracy in the color--shape task, the global
readouts reach at most 3.5\% (near the 2.8\% chance level), while the
fovea reaches 93.5\%, approaching the 99.3\% oracle. Global embeddings
shift 7 to 15 times more under label-preserving nuisance edits than under
label-changing target edits---a counterfactual nuisance-to-signal ratio
(NSR) of 7--15, where values above 1.0 mean nuisance outweighs
signal---whereas the foveated readout, like the oracle, keeps its NSR
near 0.3. On the color-free variant, the global readouts clear the 25\%
chance level but remain about 35 points below the fovea. On the
GQA-derived task, question-conditioned foveation reaches 17\% paired
accuracy, whereas the global readouts reach about 4\%, only modestly
above a 2.3\% question-only prior; an oracle given the target box
reaches 33\%.
Ultimately, our results indicate that the question ``Are frozen vision
models spatially blind?'' is under-specified: the answer depends on the
interface used to access their patch tokens.

Our contributions are threefold:
\begin{enumerate}
    \item We formulate readout-level spatial blindness and introduce a
    frozen-encoder diagnostic that distinguishes it from representation
    failure, including a counterfactual NSR computed directly on the
    embeddings.
    \item We introduce the foveated probing protocol, alongside
    controlled and natural-image evaluations with paired, localized
    tests.
    \item We demonstrate that global collapse can make available
    information appear absent, while a lightweight learned selector
    recovers most of the oracle-accessible signal.
\end{enumerate}

The remainder of the paper presents the readout protocols and the NSR
diagnostic, the construction of the three evaluation tasks, the
experimental results, and a brief conclusion. To keep the main text
focused on the diagnostic and its evidence, we defer an extensive
discussion of related work to the appendix.

%% file: 3-method.tex
\section{Method}
\label{sec:method}

\subsection{Notation and Problem Setup}
\label{sec:notation}

We study image classification via downstream readouts built on top of a frozen 
vision encoder. An example is a pair $(I, y)$ in the vision-only setting, where 
$y \in \{0, \ldots, C-1\}$, or a triple $(I, Q, y)$ in the vision--language 
setting, where $Q$ is a text question.

From image $I$, the frozen vision encoder yields final-layer patch tokens 
$X(I) = [x_1(I), \ldots, x_N(I)]^{\top} \in \mathbb{R}^{N \times d}$ and, 
optionally, a global summary token $x_{\mathrm{sum}}(I) \in \mathbb{R}^{d_{\mathrm{sum}}}$. 
For vision--language tasks, a frozen text encoder embeds $Q$ into 
$t(Q) \in \mathbb{R}^{d_t}$. A visual readout $r$ maps these representations 
to $z_r(I, Q) \in \mathbb{R}^{d_r}$ (omitting $Q$ for vision-only readouts). 
Each readout is paired with a separately trained classifier probe $h_{\phi_r}$ 
producing class logits $\ell_r$. We denote vector concatenation as $[a; b]$.

\subsection{Frozen-Encoder Readout Protocols}
\label{sec:readouts}

All visual readouts operate on patch tokens $X(I)$, except the pretrained 
summary baseline which uses $x_{\mathrm{sum}}(I)$. Vision and text encoders 
remain frozen throughout; we only train the readout parameters and probe parameters $\phi_r$.

\paragraph{Baseline Readouts.}
 The global average pooling (GAP) readout collapses patch tokens via uniform 
 averaging, $z_{\mathrm{gap}}(I) = \frac{1}{N}\sum_{i=1}^{N} x_i(I) \in \mathbb{R}^{d}$, 
 while the summary readout uses $z_{\mathrm{sum}}(I) = x_{\mathrm{sum}}(I) \in \mathbb{R}^{d_{\mathrm{sum}}}$. 
 For vision--language tasks, a question-only baseline sees only $t(Q)$, 
 measuring task performance achievable purely from text priors.

\paragraph{Foveated Readout.}
Our primary diagnostic readout is a single-query attention pool over frozen
patch tokens, derived from attention-based set pooling \cite{lee2019set} but
deliberately minimal. A query $q(Q) \in \mathbb{R}^{d}$ assigns each token a
score $s_i(I, Q) = \mathrm{LN}\left(x_i(I)\right)^{\top} q(Q) \,/\, T$,
where $\mathrm{LN}$ is feature-wise layer normalization and $T > 0$ is a
softmax temperature. Scores are normalized via
$\alpha(I, Q) = \operatorname{softmax}\left(s(I, Q)\right)$, yielding the 
weighted token vector 
$z_{\mathrm{fov}}(I, Q) = \sum_{i=1}^{N} \alpha_i(I, Q) \, x_i(I) \in \mathbb{R}^{d}$.

Layer normalization affects token selection but not the returned output space,
which remains a weighted sum of raw tokens (matching the space of GAP). For
vision-only tasks, $q$ is a single learned parameter vector shared across all
images and $T$ is fixed at the standard dot-product scale $\sqrt{d}$. For
vision--language tasks, $q(Q) = W_q \, \mathrm{LN}\left(t(Q)\right)$ with
learned linear map $W_q \in \mathbb{R}^{d \times d_t}$, and $T$ is itself
learned, initialized at $\sqrt{d}$. In both settings, the foveated
parameters ($q$ or $W_q$, and $T$ where learned) are trained jointly with
the classifier probe from the task loss alone; the attention receives no
location supervision.

\paragraph{Oracle Target-Token Readout.}
The oracle readout uses an annotation-defined target region to select patch
tokens before pooling; it is included only as a diagnostic upper bound and is
not a deployable readout. Let $m_i(I,Q)\in\{0,1\}$ indicate whether patch
token $i$ overlaps the target region. The oracle representation is the mean of the
selected patch tokens,
\begin{equation}
    z_{\mathrm{oracle}}(I,Q)
    =
    \frac{
        \sum_{i=1}^{N}
        m_i(I,Q)x_i(I)
    }{
        \sum_{i=1}^{N}
        m_i(I,Q)+\epsilon
    }
    \in\mathbb{R}^{d},
\end{equation}
where $\epsilon>0$ is a small numerical-stability constant.

\subsection{Probes}
\label{sec:probes}

Each readout $r$ is paired with a separately trained classifier probe
$h_{\phi_r}$, a small one-hidden-layer MLP; within each experiment, all
readouts share the same probe architecture and training protocol, and
results for linear probes are reported in the appendix. For vision-only
tasks, the probe receives the readout vector alone,
$\ell_r(I)=h_{\phi_r}\!\left(z_r(I)\right)\in\mathbb{R}^{C}$. For
vision--language tasks, every readout uses the same late-fusion interface,
$\ell_r(I,Q)=h_{\phi_r}\!\left(\left[z_r(I,Q);t(Q)\right]\right)$:
question-independent readouts supply the same visual vector for every
question about image $I$, so the question enters only through the
classifier, whereas the foveated and oracle readouts select patch tokens
using the question or its annotation-defined target before pooling. The 
question-only baseline receives $t(Q)$
alone.

\subsection{Counterfactual Nuisance-to-Signal Ratio for Synthetic Tasks}
\label{sec:synthetic_nsr}

A useful readout should change when the content that determines the label
changes, and remain stable under irrelevant scene variation. We measure these 
two properties---signal and nuisance---directly
on the readout representations. Because readouts differ in dimension and scale,
all NSR quantities use the standardized vector
$\widetilde{z}_r(I)\in\mathbb{R}^{d_r}$, obtained by z-scoring each
coordinate of $z_r(I)$ on the base training split.

\input{figures_tex/nsr_example}

\paragraph{Signal and nuisance variation.}
Both quantities are computed on held-out diagnostic collections, indexed by
a nuisance regime $b\in\mathcal{B}$. The first collection,
$\mathcal{P}^{\mathrm{syn}}_b$, contains counterfactual pairs
$(I,I^{\mathrm{cf}})$ in which the label-defining factor is edited while the
nuisance variables of regime $b$ are held fixed; we define the label-changing
signal preserved by readout $r$ as the mean squared per-coordinate change
across such pairs,
\begin{equation}
    S^{\mathrm{syn}}_r(b)
    =
    \mathbb{E}_{
        (I,I^{\mathrm{cf}})
        \sim
        \mathcal{P}^{\mathrm{syn}}_b
    }
    \left[
        \frac{1}{d_r}
        \left\|
            \widetilde{z}_r(I^{\mathrm{cf}})
            -
            \widetilde{z}_r(I)
        \right\|_2^2
    \right].
\end{equation}
The second collection,
$\mathcal{G}^{\mathrm{syn}}_b$, contains label-preserving nuisance groups
$G=\{I^{(1)},\ldots,I^{(|G|)}\}$ whose images share the label-defining factor
but differ in nuisance content; we define nuisance variation as the mean
per-coordinate variance within a group,
\begin{equation}
    N^{\mathrm{syn}}_r(b)
    =
    \mathbb{E}_{
        G\sim\mathcal{G}^{\mathrm{syn}}_b
    }
    \left[
        \frac{1}{d_r}
        \sum_{j=1}^{d_r}
        \operatorname{Var}_{I\in G}
        \left[
            \widetilde{z}_{r,j}(I)
        \right]
    \right].
\end{equation}

The counterfactual nuisance-to-signal ratio is
\begin{equation}
    \label{eq:nsr_syn}
    \mathrm{NSR}^{\mathrm{syn}}_r(b)
    =
    \frac{
        N^{\mathrm{syn}}_r(b)
    }{
        S^{\mathrm{syn}}_r(b)
        +
        \varepsilon_{\mathrm{NSR}}
    },
\end{equation}
where $\varepsilon_{\mathrm{NSR}}>0$ is a small numerical-stability constant.
A low NSR is desirable: the representation then varies mainly with the label
rather than with irrelevant scene content (Figure~\ref{fig:cf_nuisance_example}).

\paragraph{Signal-dilution interpretation.}
To illustrate why global pooling can produce a high NSR, consider one
counterfactual pair $(I,I^{\mathrm{cf}})$ in the original, unstandardized
token space. Let
$\mathcal{I}_{\mathrm{tar}}\subseteq\{1,\ldots,N\}$
be the set of patch-token indices affected by the target edit, and let
$M=|\mathcal{I}_{\mathrm{tar}}|$
be the number of affected tokens. For each patch token, define its
counterfactual displacement as
$\Delta x_i = x_i(I^{\mathrm{cf}}) - x_i(I)$,
and define the mean displacement over the target-token set as
$\Delta\overline{x}_{\mathrm{tar}}
= \frac{1}{M}\sum_{i\in\mathcal{I}_{\mathrm{tar}}}\Delta x_i$.
If token displacements outside the target region are negligible, the change in
the GAP representation satisfies
\begin{align}
    \Delta z_{\mathrm{gap}}
    &\coloneqq
    z_{\mathrm{gap}}(I^{\mathrm{cf}})
    -
    z_{\mathrm{gap}}(I)
    \\
    &=
    \frac{1}{N}
    \sum_{i=1}^{N}
    \Delta x_i
    \approx
    \frac{M}{N}
    \Delta\overline{x}_{\mathrm{tar}}.
\end{align}
Thus, before coordinate standardization, global averaging attenuates the
localized representation shift in proportion to the fraction $M/N$ of affected
tokens. Because $S^{\mathrm{syn}}_r(b)$ is based on squared distances, the
corresponding signal contribution can be attenuated approximately in proportion
to $(M/N)^2$.

For the foveated readout, suppose that its attention weights are
approximately stable across the counterfactual pair. The representation change 
is approximately
\begin{equation}
    \Delta z_{\mathrm{fov}}
    \coloneqq
    z_{\mathrm{fov}}(I^{\mathrm{cf}})
    -
    z_{\mathrm{fov}}(I)
    \approx
    \sum_{i\in\mathcal{I}_{\mathrm{tar}}}
    \alpha_i(I)\,
    \Delta x_i.
\end{equation}
Define the attention mass assigned to the affected target tokens as
$\omega_{\mathrm{tar}}(I)
= \sum_{i\in\mathcal{I}_{\mathrm{tar}}}\alpha_i(I)$.
When the target-token displacements are similarly oriented, the foveated
shift scales with $\omega_{\mathrm{tar}}(I)$ rather than $M/N$: concentrating
attention on the target preserves the localized label-changing signal that
uniform pooling dilutes, and excludes the irrelevant distractor variation
that uniform pooling passes into $N^{\mathrm{syn}}_r(b)$; the two effects
compound into a larger NSR for the global readouts.

%% file: figures_tex/nsr_example.tex
\begin{figure}[t]
  \centering
  \includegraphics[width=\linewidth]{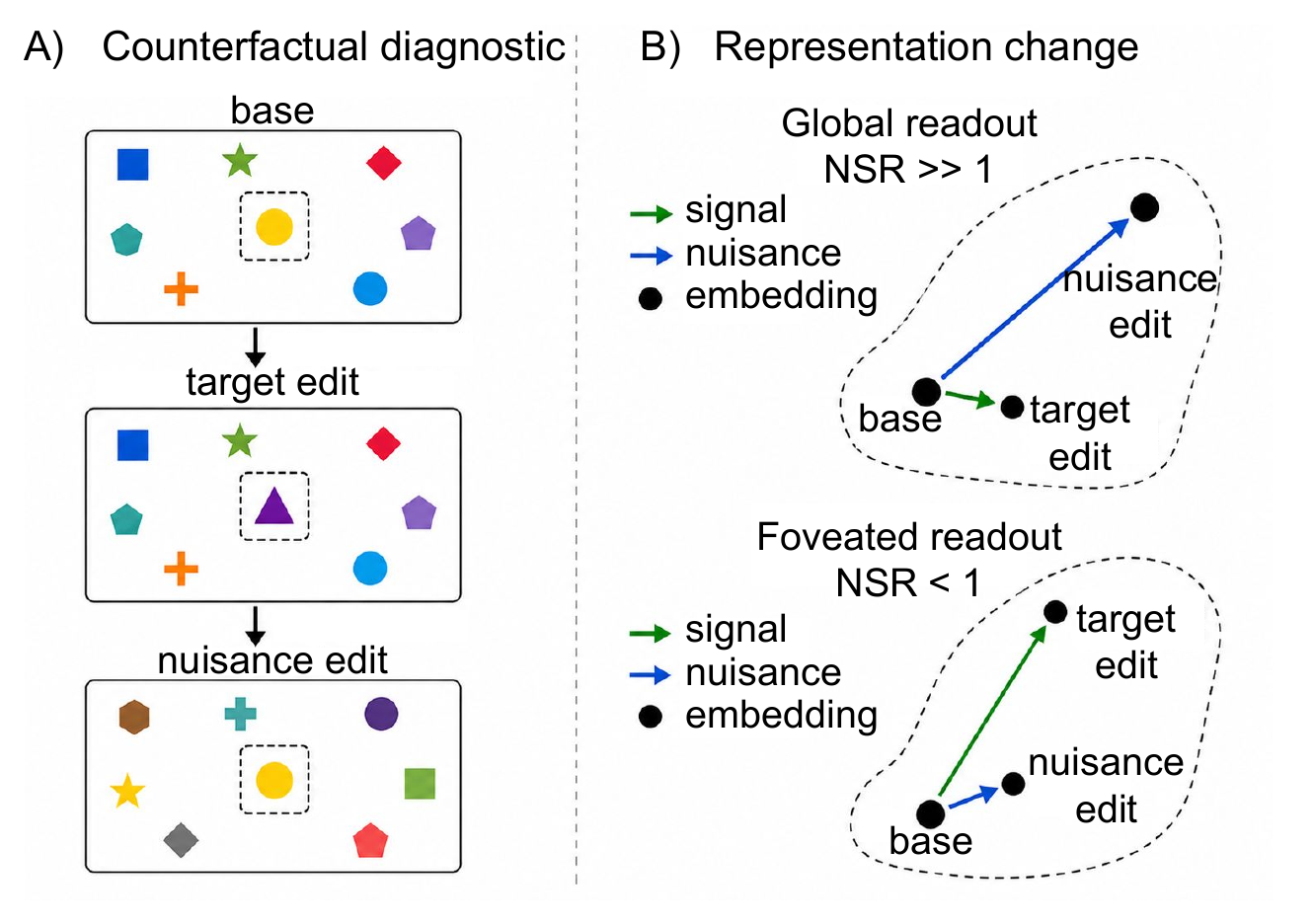}
\caption{\textbf{NSR construction.}
(A) The counterfactual changes the target (dashed box) while holding the
background fixed, whereas the nuisance edit changes the background while
holding the target fixed. (B) A good readout exhibits a large representation
change under the label-changing counterfactual (high signal) but only a small
change under the nuisance edit (low nuisance), yielding $\mathrm{NSR}<1$, as
for the foveated readout. Global readouts show the opposite pattern---small
signal and large nuisance---yielding $\mathrm{NSR}\gg 1$.}
  \label{fig:cf_nuisance_example}
\end{figure}

%% file: 4-tasks_and_datasets.tex
\section{Tasks and Datasets}
\label{sec:evaluation_tasks}

We evaluate the frozen-encoder readouts on three localized binding tasks. The
first is a controlled vision-only task in which the label is the color--shape
conjunction of one target object embedded among adversarial distractors; its
synthetic construction permits label-changing counterfactuals and
label-preserving nuisance groups. The second is a color-free crowded
shape-detection variant of the first. The third is a vision--language task
constructed from natural images, in which two questions about the same image
select different same-category object instances with different colors.
Together, the tasks test whether localized binding information is absent from
the frozen representation or merely poorly exposed by a global readout.
Additional construction details for the datasets are given in the appendix.

\subsection{Synthetic Binding Tasks under Clutter}
\label{sec:color_binding_task}

Classical accounts of visual binding motivate this task: feature-integration
theory proposes that conjunctions of separable attributes, such as color and
shape, require focused attention, while object-file theory emphasizes the
integration of features within object-specific representations
\cite{treisman1980feature,kahneman1992reviewing}. We therefore construct the
color--shape binding task (CSB), a six-way vision-only classification task in
which each image is a $384\times384$ white canvas containing one target
object, rendered at $14$\,px, and
$K\sim\operatorname{Uniform}\{10,\ldots,50\}$ distractors
(Figure~\ref{fig:syn_examples}, top). The target's color and shape
are sampled uniformly from
$c\in\{\texttt{red},\texttt{yellow},\texttt{purple}\}$ and
$s\in\{\texttt{circle},\texttt{triangle}\}$, and the label is the
conjunction $(c,s)$, giving $C=6$ classes. The target triangle is upright;
rotated triangles occur only as distractor shapes. The remaining
palette colors \texttt{orange}, \texttt{green}, and \texttt{blue} are
distractor-only, and non-target distractor shapes are sampled from an
$18$-shape vocabulary. We generate $10{,}000$ base images, divided
$6{,}000/2{,}000/2{,}000$ into training, validation, and test splits with the
six conjunctions sampled uniformly in every split, plus a clean
evaluation-only split of $500$ images in which the target appears alone
($K=0$).

\input{figures_tex/syn_examples}

\paragraph{Adversarial lures.}
Each scene contains one to three lures of each of three types. A
\emph{target-shape lure} has the target shape but a distractor-only color; a
\emph{counterpart-shape lure} has the other target shape, also with a
distractor-only color; and a \emph{color lure} has the target color but a
non-target shape. We collect the three per-type counts into the lure-count
tuple $\lambda$. The remaining distractors are noise objects whose shapes are
sampled from the non-target shape vocabulary and whose colors are sampled from
the complete six-color palette. Every scene therefore contains irrelevant
objects that match the target color or the target shape, while only the target
object contains the label-defining conjunction.

\paragraph{Counterfactual pairs and nuisance groups.}
From the base test scenes, we construct evaluation-only diagnostic images of
two kinds: $9{,}366$ label-changing counterfactual pairs, in which the target
color and/or shape is swapped with a matched non-target object while
preserving object centers, count, scale, and the marginal color and shape
histograms; and $16$ label-preserving nuisance variants per scene
($32{,}000$ images), which resample non-target objects while fixing the
target, $K$, and $\lambda$. Assigned to nuisance regimes
$b\in\mathcal{B}$ by the distractor-count bins
$\{10\text{--}19, 20\text{--}29, 30\text{--}39, 40{+}\}$, these form the
collections $\mathcal{P}^{\mathrm{syn}}_b$ and $\mathcal{G}^{\mathrm{syn}}_b$.

\paragraph{Crowded shape detection (CSD).}
To test shape binding in the absence of color cues, we additionally construct
a color-free variant of the synthetic task
(Figure~\ref{fig:syn_examples}, bottom). Scenes are generated as in CSB
except that all objects are rendered in black, no adversarial lures are used,
and distractor sizes are sampled from $1\times$ to $11.5\times$ the target
size, with the target rendered at $7$\,px, half its CSB size. The target is
the scene's unique object with a target shape; its shape is the label,
$s\in\{\texttt{circle},\texttt{triangle}\}$ ($C=2$). Distractor shapes are
drawn from the same non-target vocabulary as in CSB. We use the same split
sizes as CSB, and each base test image is paired with one label-changing
counterfactual in which the target shape is flipped in place. On CSD we
report only paired-counterfactual accuracy; it serves as a color-free
control for the accuracy comparison, and the representation-level NSR
analysis is conducted on CSB.

\subsection{GQA-Derived Localized Color Binding}
\label{sec:gqa_color_binding_dataset}

\input{figures_tex/task_examples}

To test localized color binding in natural images, we construct a paired
vision--language task from GQA scene graphs and Visual Genome images
\cite{hudson2019gqa,krishna2017visual}, generating questions directly from
scene-graph object annotations rather than using the original GQA questions.
We refer to the resulting task as GQA$^{*}$.
Each pair contains two examples derived from the same image: the two
questions name the same object category but select different object instances
with different annotated colors, following the template
``What color is the \{relation\} \{name\}?'' with a frame-relative relation
in \{\texttt{left}, \texttt{right}, \texttt{above}, \texttt{below}\}
(Figure~\ref{fig:gqa_pair_example}). Because the two target
instances have different color labels, answering both questions requires
selecting the object specified by the question rather than assigning one
global color label to the image. Restricting the answer vocabulary to the
ten most frequent colors gives a classification task with $C=10$ and
naturally imbalanced labels; the dataset contains $22{,}502$ question
examples
($11{,}251$ same-image pairs over $9{,}950$ images, spanning $542$ target
object categories), divided into image-disjoint training, validation, and
test splits.

%% file: figures_tex/syn_examples.tex
\begin{figure}[t]
  \centering
  \includegraphics[width=\linewidth]{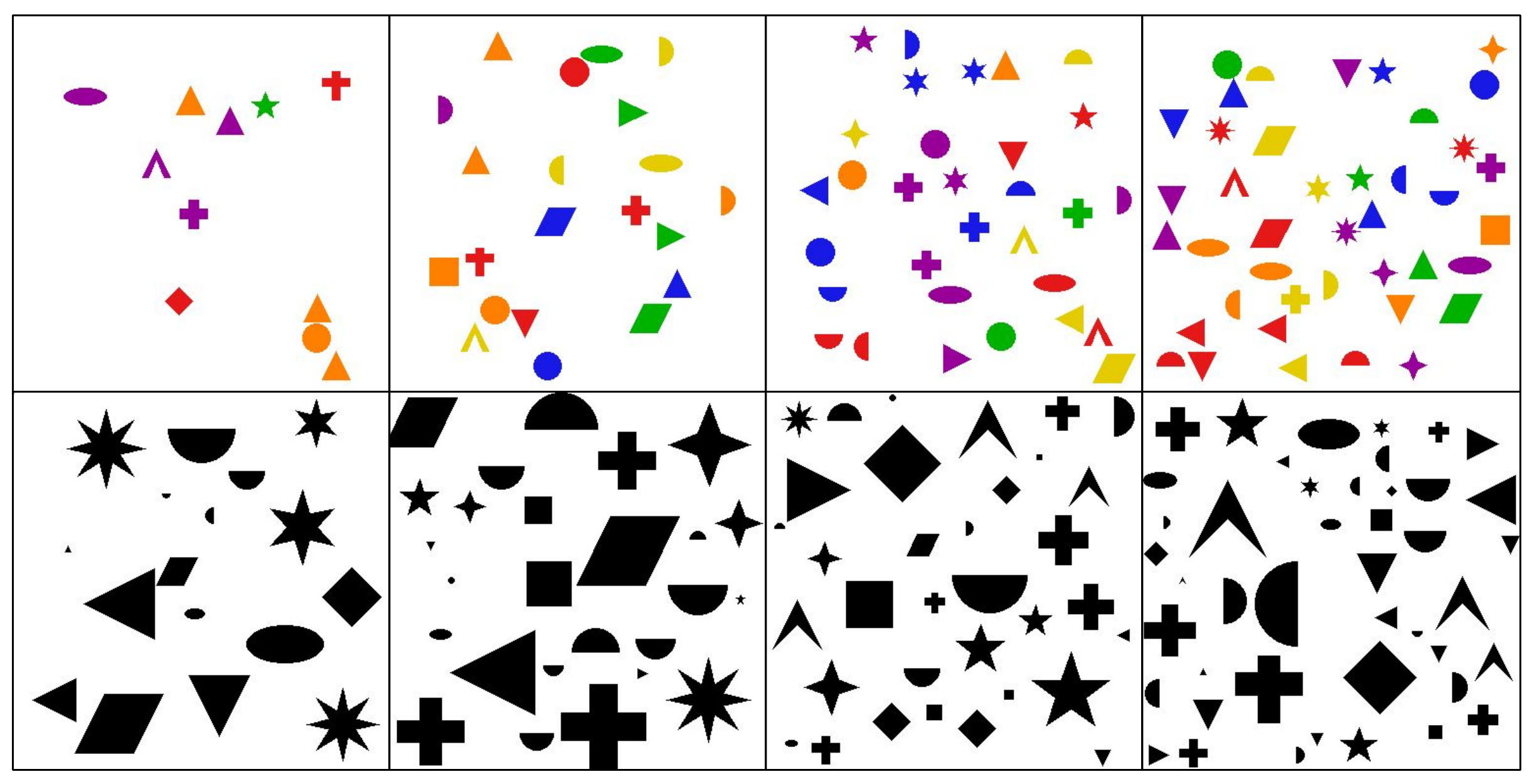}
  \caption{Example scenes from the two synthetic tasks at increasing
  distractor count $K$, one per distractor-count bin. \emph{Top}:
  color--shape binding (CSB); each scene contains a single target object
  whose color--shape conjunction defines the six-way label, among
  distractors that separately match its color or shape. \emph{Bottom}:
  crowded shape detection (CSD), a color-free variant; all objects are
  black and the label is the shape of the scene's unique target-shape
  object.}
  \label{fig:syn_examples}
\end{figure}

%% file: figures_tex/task_examples.tex
\begin{figure}[t]
  \centering
    \includegraphics[width=0.9\linewidth]{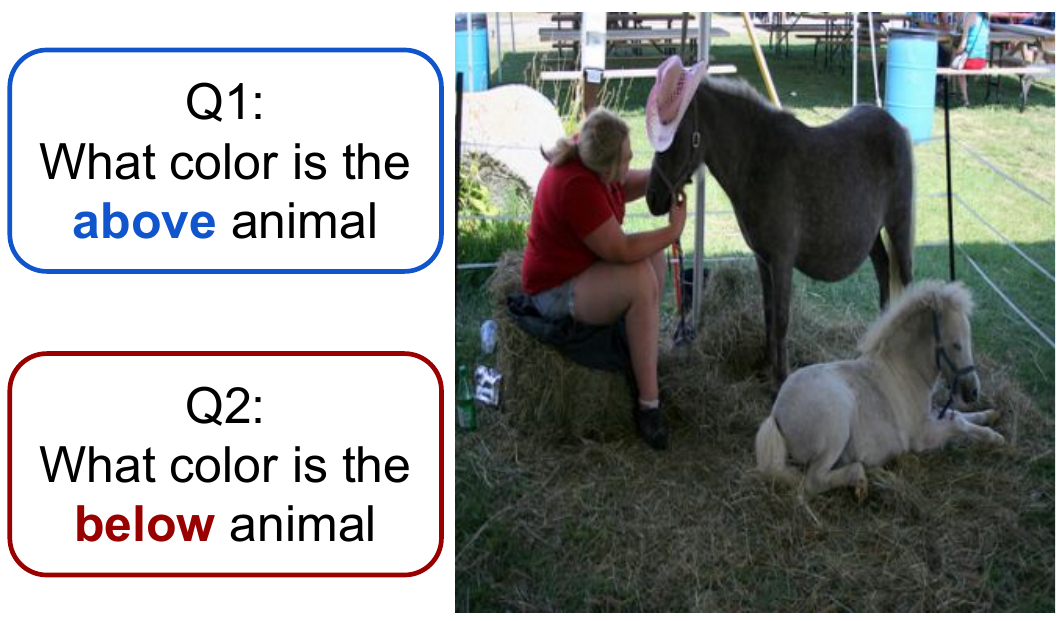}
  \caption{Example pair from the GQA-derived localized color binding task.
  Both questions are asked about the same image and name the same category,
  but select different instances: ``What color is the \emph{above} animal?''
  (\texttt{brown}) and ``What color is the \emph{below} animal?''
  (\texttt{white}). A readout that assigns one global color to the image
  cannot answer both questions correctly.}
  \label{fig:gqa_pair_example}
\end{figure}

%% file: 5-experiments.tex
\section{Experiments}
\label{sec:experiments}

We evaluate the readout interfaces on two settings: controlled synthetic tasks and a GQA-derived task on natural images. In both, a frozen SigLIP vision encoder is held fixed and only the readout and its probe are trained, so differences in performance reflect how each interface exposes the information already present in the patch tokens rather than differences in the encoder. We compare question-independent global readouts, the pretrained summary representation and global average pooling (GAP), against a question-conditioned foveated readout and an oracle readout that pools the ground-truth target region. Beyond ordinary classification accuracy, we use matched counterfactuals that isolate sensitivity to the label-defining target conjunction, allowing us to separate classification performance from sensitivity to the target binding. We first evaluate the readouts on the controlled synthetic task, where nuisance content can be increased in a controlled way and its effect measured through a counterfactual nuisance-to-signal analysis, and then test whether the readout-level distinction extends to natural images.

\subsection{Experimental Protocol}
\label{sec:experimental_protocol}

All experiments use the same frozen SigLIP vision encoder
\cite{zhai2023sigmoid}, \texttt{siglip-so400m-patch14-384}, at its native
\(384\times384\) resolution, from which we extract the final-layer
\(N=729\) patch tokens of dimension \(d=1152\), plus the pretrained summary
representation when a readout requires it; no gradients flow through the
encoder. Every readout is paired with a separately trained classifier
probe, and only the trainable readout (where applicable) and the probe are
optimized. Each readout--probe configuration is trained with five seeds,
and metrics are reported as mean \(\pm\) standard deviation. Results for
additional frozen encoders, CLIP \cite{radford2021learning} and SigLIP~2
\cite{tschannen2025siglip}, are reported in the appendix and show the same
trends. Task-specific training protocols are provided in the appendix.

\subsubsection{Synthetic Evaluation Measures}
\label{sec:synthetic_protocol}

For a trained readout \(r\), let \(\hat{y}_r(I)\) denote its predicted class.
The base test set has \(n_{\mathrm{test}}\) examples \((I_i,y_i)\).
For each base test image \(I_i\), let \(\mathcal{J}_i\) index its
label-changing counterfactuals. We write \(I^{\mathrm{cf}}_{ij}\) for
counterfactual \(j\in\mathcal{J}_i\) and
\(y^{\mathrm{cf}}_{ij}\neq y_i\) for its new label. The set of evaluated
counterfactual pairs is
\(\mathcal{C}^{\mathrm{syn}}_{\mathrm{test}}=\{(i,j):1\leq i\leq n_{\mathrm{test}},\ j\in\mathcal{J}_i\}\).
For brevity, we write
\(c_r(i)=\mathbf{1}[\hat{y}_r(I_i)=y_i]\)
and
\(c^{\mathrm{cf}}_r(i,j)=\mathbf{1}[\hat{y}_r(I^{\mathrm{cf}}_{ij})=y^{\mathrm{cf}}_{ij}]\)
for correct predictions on base images and counterfactuals.
Base test accuracy is
\begin{equation}
    \mathrm{Acc}_{\mathrm{test}}(r)
    =
    \frac{1}{n_{\mathrm{test}}}
    \sum_{i=1}^{n_{\mathrm{test}}}
    c_r(i).
\end{equation}
Paired-counterfactual accuracy requires both the base image and its
counterfactual to be classified correctly:
\begin{equation}
    \mathrm{Acc}_{\mathrm{paired\text{-}cf}}(r)
    =
    \frac{
        1
    }{
        \left|
            \mathcal{C}^{\mathrm{syn}}_{\mathrm{test}}
        \right|
    }
    \sum_{
        (i,j)\in
        \mathcal{C}^{\mathrm{syn}}_{\mathrm{test}}
    }
    c_r(i)\,
    c^{\mathrm{cf}}_r(i,j).
\end{equation}
If a base image has multiple counterfactuals, it contributes once for
each pair. Under independent uniform guessing, the expected
paired-counterfactual accuracy is
\(\mathbb{E}[\mathrm{Acc}_{\mathrm{paired\text{-}cf}}]=1/C^{2}\):
approximately \(2.8\%\) for CSB (\(C=6\)) and \(25\%\) for CSD (\(C=2\)).

\paragraph{Nuisance-to-signal analysis.}
We compute the counterfactual nuisance-to-signal ratio of
Equation~\eqref{eq:nsr_syn} using the pre-probe representations. Each readout
coordinate is standardized using its mean and standard deviation on the base
training split. Within each distractor-count bin \(b\),
\(S_r^{\mathrm{syn}}(b)\) is computed from the label-changing
counterfactual pairs, and \(N_r^{\mathrm{syn}}(b)\) is computed from the
label-preserving nuisance groups. We set
\(\varepsilon_{\mathrm{NSR}}=10^{-8}\).

\subsubsection{GQA$^{*}$ Evaluation Measures}
\label{sec:gqa_protocol}

For a trained readout \(r\), let \(\hat{y}_r(I,Q)\) denote its predicted color
for image--question pair \((I,Q)\).
The test split contains \(n_{\mathrm{pair}}\) same-image pairs, for a total of
\(2n_{\mathrm{pair}}\) question examples. Pair
\(p\in\{1,\dots,n_{\mathrm{pair}}\}\) consists of an image \(I_p\) together
with two question--label pairs \((Q_{p,1},y_{p,1})\) and
\((Q_{p,2},y_{p,2})\) that refer to different object instances of the same
category with different colors, so \(y_{p,1}\neq y_{p,2}\). Distinct pairs can
share the same image. Analogously to the synthetic task, we write
\(c_r(p,j)=\mathbf{1}[\hat{y}_r(I_p,Q_{p,j})=y_{p,j}]\) for a correct
prediction.
Our primary localized-binding measure is paired accuracy,
\begin{equation}
    \mathrm{Acc}_{\mathrm{paired}}(r)
    =
    \frac{1}{n_{\mathrm{pair}}}
    \sum_{p=1}^{n_{\mathrm{pair}}}
    c_r(p,1)\,
    c_r(p,2),
\end{equation}
which credits a pair only when both localized selections on the shared image
are correct.
Under independent uniform guessing over the \(C=10\) color labels, the expected
paired accuracy is
\(\mathbb{E}[\mathrm{Acc}_{\mathrm{paired}}]=1/C^2=0.01\).

\subsection{Global Readout Failure under Synthetic Clutter}
\label{sec:exp_csb}

We first evaluate the readout interfaces on the controlled synthetic tasks.
The synthetic construction allows us to distinguish ordinary
classification performance from sensitivity to the label-defining target
conjunction, and to measure how both change as nuisance content increases.

\subsubsection{Main Readout Comparison and Counterfactual Sensitivity}
\label{sec:csb_main_results}

\input{figures_tex/paired_acc_bars}

All readouts are near ceiling (\(\geq 99.6\%\)) on the clean no-distractor
split, so the frozen encoder represents the isolated colors and shapes needed
for the task; the full per-readout table is provided in the appendix. Under
clutter, the matched counterfactual evaluation separates the readouts sharply
(Figure~\ref{fig:binding-main}): the global readouts obtain
paired-counterfactual accuracies of \(3.2 \pm 0.1\%\) (Summary) and
\(3.5 \pm 0.2\%\) (GAP) against the \(2.8\%\) chance level, whereas the
foveated readout reaches \(93.5 \pm 0.6\%\), within \(5.8\) points of the
\(99.3\%\) oracle upper bound versus a roughly \(96\)-point gap for the
global readouts. On the color-free CSD variant, which removes color cues and
adversarial lures, the global readouts clear the \(25\%\) chance level
(Summary \(53.6 \pm 0.4\%\), GAP \(58.9 \pm 0.4\%\)) but remain about
\(35\) points below the foveated readout (\(93.7 \pm 1.0\%\)), which again
approaches the oracle (\(97.9 \pm 0.1\%\)). Most of the target-binding
signal is thus present in the frozen patch tokens but is poorly exposed by
global spatial collapse.

\subsubsection{Clutter Scaling and Nuisance-to-Signal Analysis}
\label{sec:csb_clutter_results}
\label{sec:csb_nsr_results}

\input{figures_tex/csb_nsr}

Figure~\ref{fig:acc_vs_nsr} relates base-test accuracy to the counterfactual
NSR by distractor-count bin. Although all readouts are near ceiling on the
clean split, the global readouts deteriorate steadily with clutter: from the
\(K=10\text{--}19\) bin to the \(K=40+\) bin, GAP loses \(20.9\) points of
accuracy while its NSR grows from \(6.97\) to \(15.40\), and the pretrained
summary readout behaves similarly. The foveated readout instead loses only
\(3.1\) points and keeps its NSR near \(0.30\) across all bins, comparable to
the oracle (NSR \(0.29\text{--}0.40\)). This matches the signal-dilution
account behind the counterfactual NSR: global pooling attenuates the
localized representation change induced by the target edit while continuing to
expose the probe to nuisance variation, whereas selective pooling preserves
the label-changing signal.

\subsubsection{Adversarial-Lure Analysis}
\label{sec:csb_lure_results}

\input{figures_tex/csb_lure}

Figure~\ref{fig:acc_by_lure} examines sensitivity to the composition of the
adversarial lures. The global readouts depend strongly on it: going from one
to three lures raises GAP accuracy by \(28.4\) points for target-shape lures,
lowers it by \(24.9\) points for counterpart-shape lures, and raises it by
\(15.3\) points for color lures, with the pretrained summary readout behaving
similarly. This is consistent with the global probes counting scene-level
features: color and target-shape lures help, counterpart-shape lures hurt. The
foveated readout is almost invariant to these manipulations
(\(97.2\text{--}98.6\%\)), as is the oracle (\(99.3\text{--}99.9\%\)).

\subsection{Question-Conditioned Foveation on Natural Images}
\label{sec:exp_gqa_color_binding}

We next test whether the readout-level distinction observed in the controlled
synthetic tasks extends to natural images. In the GQA-derived task, the two
examples in each pair share the same image and object-category name but ask
about different same-category object instances with different colors.

\subsubsection{Main Readout Comparison}
\label{sec:gqa_main_results}

Figure~\ref{fig:binding-main} reports the main results. The question-only
baseline reaches \(2.3 \pm 0.4\%\) paired accuracy: above the \(1\%\) chance
level, so the generated questions and object-category names carry some color
prior, but one far too weak to answer both localized questions about the same
image correctly. Adding a question-independent global visual
representation improves paired accuracy only modestly (Summary
\(3.8 \pm 0.6\%\), GAP \(4.3 \pm 0.3\%\)); although the late-fusion classifier
receives a different question for each example, both examples in a pair
receive the same visual vector after question-independent spatial collapse.

Question-conditioned foveation is substantially stronger, reaching
\(17.0 \pm 0.7\%\) paired accuracy---about
\(4\times\) the global readouts---because the question constructs the
attention query, allowing the two questions in a same-image pair to select
different visual evidence before the patch tokens are collapsed into a single
vector. The oracle readout reaches \(33.0 \pm 0.7\%\) paired accuracy;
because it uses the target box only to pool frozen patch tokens and is not
given the color label, the frozen spatial representation contains additional
object-specific color information that the learned fovea does not yet recover.
The oracle's absolute level largely reflects the benchmark rather than the
representation: scene-graph bounding boxes and color labels are noisy, color
naming is often genuinely ambiguous (many objects are multi-colored), and
paired accuracy compounds the errors of two questions.
Still, foveation closes roughly \(45\%\) of the summary-to-oracle gap in
paired accuracy. The resulting ordering
(question only \(<\) summary/GAP \(<\) foveated \(<\) oracle) supports the
same interface-level interpretation as the synthetic experiment: substantially
more localized information is available in the frozen patch tokens than is
exposed by a question-independent global image representation.

\subsubsection{Attention Concentration Diagnostics}
\label{sec:gqa_attention_diagnostics}

We report attention statistics for
the learned-temperature foveated readout on the validation split at the
validation-selected checkpoint. Let \(\alpha_{ij}\) be the attention weight
assigned to patch \(j\) for validation example \(i\), and let
\(H_i(\alpha)=-\sum_{j=1}^{N}\alpha_{ij}\log\alpha_{ij}\) be its attention
entropy. We report the effective number of attended patches,
\(\mathrm{EffPatches}
=\frac{1}{n_{\mathrm{val}}}\sum_{i=1}^{n_{\mathrm{val}}}\exp(H_i(\alpha))\),
and the mean maximum attention weight,
\(\mathrm{MaxAttn}
=\frac{1}{n_{\mathrm{val}}}\sum_{i=1}^{n_{\mathrm{val}}}\max_{1\leq j\leq N}\alpha_{ij}\).
\(\mathrm{EffPatches}\) equals \(1\) for one-hot attention and \(N=729\) for
uniform attention. The reported diagnostics are averaged first over
validation examples and then over seeds.

The attention temperature is initialized at
the standard dot-product scale, \(T_0=\sqrt{d}\approx 33.9\),
but the selected models converge to a temperature of
\(14.2 \pm 3.3\). Because the attention logits are divided by the temperature
before the softmax, the lower learned value produces a sharper attention
distribution than the initialization.

The learned attention remains soft rather than collapsing to a hard
single-patch selection. Its effective number of attended patches is
\(5.2 \pm 1.2\), and its mean maximum patch weight is
\(0.621 \pm 0.035\). Thus, the
foveated readout concentrates most of its attention on a small subset of the
\(729\) patch tokens rather than approaching uniform global pooling.

%% file: figures_tex/paired_acc_bars.tex
\begin{figure}[t]
  \centering
  \includegraphics[width=\linewidth]{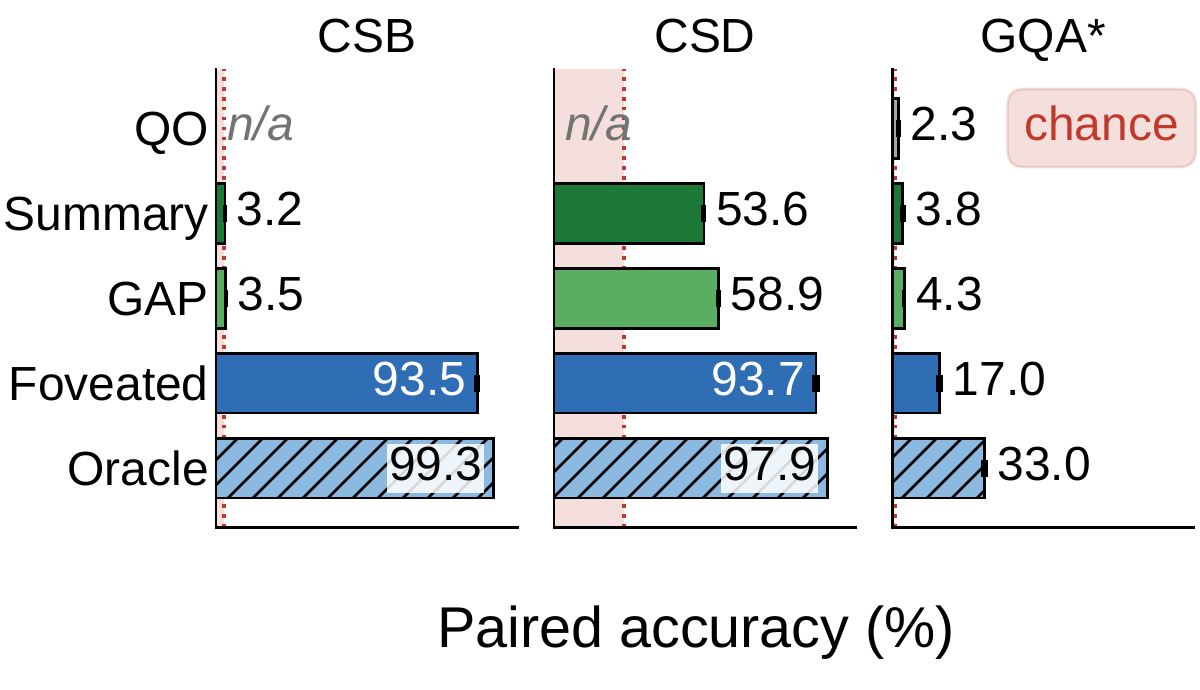}
  \caption{Paired accuracy (\%, mean $\pm$ std over seeds) on color--shape
  binding (CSB), crowded shape detection (CSD), and the GQA-derived
  localized color-binding task (GQA$^{*}$). QO (question only) is a
  language-only baseline, Summary and GAP are pooled visual readouts,
  Foveated is the learned attention-pooling readout, and the hatched
  Oracle pools frozen patch tokens inside the target region as an upper
  bound. Red bands mark the per-task chance levels ($2.8\%$, $25\%$, and
  $1\%$); ``n/a'' marks conditions that do not apply.}
  \label{fig:binding-main}
\end{figure}


%% file: figures_tex/csb_nsr.tex
\begin{figure}[t]
  \centering
  \includegraphics[width=0.99\linewidth]{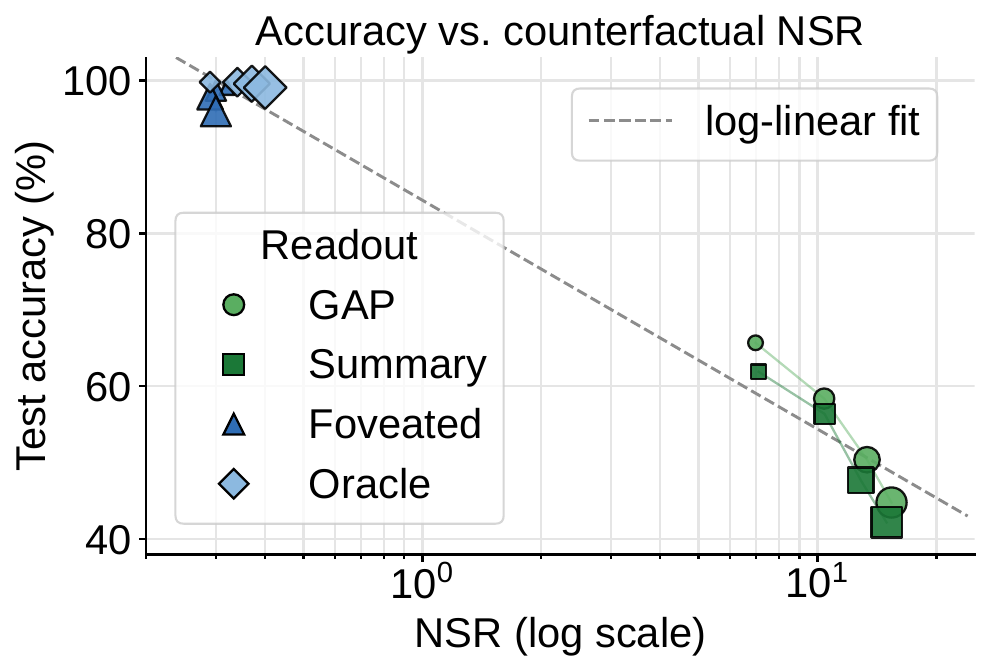}
  \caption{Base test accuracy versus counterfactual nuisance-to-signal ratio
  (NSR) by distractor-count bin \(K\), MLP probe. Each point is one
  readout evaluated in one bin (the \(K=0\) bin is omitted, as with no
  distractors there is no nuisance); marker size grows with \(K\). The pooled
  readouts (GAP, Summary) occupy the high-NSR, low-accuracy region, and their
  accuracy falls monotonically as clutter---and NSR---increase. The spatially
  selective readouts (Foveated, Oracle) remain in the low-NSR, high-accuracy
  corner across all bins. The dashed line is a log-linear fit across all
  points, indicating that higher NSR is associated with lower accuracy.}
  \label{fig:acc_vs_nsr}
\end{figure}


%% file: figures_tex/csb_lure.tex
\begin{figure}[t]
  \centering
  \includegraphics[width=\linewidth]{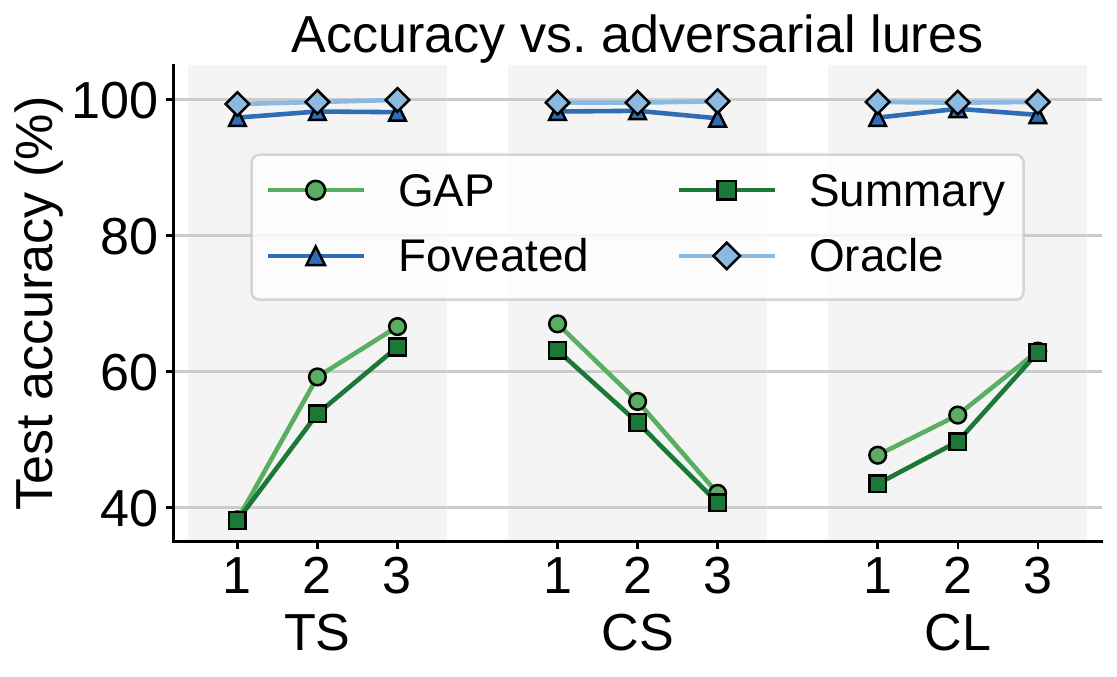}
  \caption{Accuracy under adversarial lures (MLP probe). Blocks show
  target-shape (TS), counterpart-shape (CS), and color (CL) lures; within
  each block the x-axis is the lure count (\(1\text{--}3\)). The global
  readouts (GAP, Summary) swing strongly with lure composition---rising with target-shape and color
  lures, falling with counterpart-shape lures---while the foveated and oracle
  readouts stay near ceiling throughout.}
  \label{fig:acc_by_lure}
\end{figure}

%

%% file: 6-conclusion.tex
\section{Conclusion}
\label{sec:conclusion}

Failures on spatial and compositional tasks are often taken as evidence that
vision foundation models do not represent the required information. Our
results show that this conclusion can depend critically on the interface used
to inspect the model. Holding a vision encoder fixed and varying only its
readout, we find that localized binding information can remain available in
the final patch tokens while becoming difficult to recover after those tokens
are compressed into a question-independent global image vector. Poor
performance from a global embedding therefore does not, by itself, establish
a representational absence.

The controlled experiments make this distinction explicit. On color--shape
binding, global readouts deteriorate steadily as clutter grows and are
largely insensitive to matched edits that change the target label while
preserving global color and shape statistics, whereas a single-query foveated
readout, trained without any location supervision, reaches near-oracle
accuracy and remains nearly invariant to clutter and lure composition; the
crowded shape-detection variant shows that the gap persists even without
color cues. The counterfactual nuisance-to-signal ratio supplies the
mechanism: global pooling attenuates the localized change that determines the
label while continuing to expose the probe to variation from irrelevant
objects. The natural-image experiment shows that the same issue is not
confined to synthetic scenes: letting the question construct the attention
query before spatial collapse yields roughly four times the paired accuracy
of the global readouts.

Our experiments span three frozen vision--language encoders and focus on
localized attribute binding; extending the diagnostic to other architectures,
layers, pretraining objectives, and more complex relational tasks is future
work. The foveated probe is likewise not a replacement for architectures with
richer cross-modal interaction: its value is that it changes the readout
while leaving the underlying representation fixed, directly testing where
task information becomes inaccessible. These findings suggest a broader
evaluation principle: before attributing a localized failure to the
representation learned by a foundation model, one should test whether the
conclusion survives a readout that preserves or selectively accesses its
spatial tokens. Global embeddings are useful summaries, but they should not
be mistaken for complete accounts of what a frozen vision model knows.

%% file: appendix.tex

\appendix

\section{Technical Appendix \& Supp. Material}

In this appendix we provide the material deferred from the main paper. We
begin with an extended discussion of related work, then report the full
result tables promised in the main text: the complete readout comparison,
the accuracy-versus-NSR analysis, and the per-lure accuracy breakdown,
each extended from the main paper's SigLIP-with-MLP-probe configuration to
the CLIP and SigLIP~2 encoders and to linear probes. We then document the
task-specific training protocols, the construction of the three evaluation
datasets---the color--shape binding task (CSB), the crowded shape-detection
task (CSD), and the GQA-derived localized color-binding task
(GQA$^{*}$)---and the computing infrastructure.

\input{2-related}

\subsection{Additional Results}

This subsection reports the full result tables promised in the main
paper. Each table extends one main-paper analysis---the readout
comparison, the accuracy-versus-NSR analysis, and the per-lure
accuracy breakdown---from the SigLIP encoder with an MLP probe to all
three encoders and both probe heads. We briefly discuss each table in
turn.

\subsubsection{Full Readout Comparison}

Table~\ref{tab:app_promised} reports clean and paired accuracy for every
combination of encoder (SigLIP, CLIP, SigLIP~2), probe (linear, MLP), and
readout. The ordering established in the main paper replicates in every
configuration: on CSB, the global readouts (Summary, GAP) reach at most
$8.7\%$ paired-counterfactual accuracy---near the $2.8\%$ chance
level---despite near-perfect clean accuracy in most configurations, while
the foveated readout reaches $93.5$--$99.2\%$, close to the oracle. The
separation persists on the color-free CSD task and on GQA$^{*}$, where
foveation improves paired accuracy approximately $4\times$ or more over
the global readouts for every encoder. The gap is not a matter of probe
capacity: linear probes show the same pattern as MLPs.

\subsubsection{Accuracy versus NSR under Clutter}

Table~\ref{tab:app_acc_vs_nsr} extends the accuracy-versus-NSR analysis of
the main paper to all encoders and probes. For every configuration, the
global readouts pair falling accuracy with an NSR that grows from
approximately $5$--$8$ in the $K{=}10$--$19$ bin to $9$--$17$ at
$K{=}40{+}$: nuisance edits move the embedding far more than
label-changing ones, and increasingly so under clutter. The foveated
and oracle readouts instead hold NSR approximately constant
($0.16$--$0.40$) with accuracy of $96\%$ or higher in every bin. 

\subsubsection{Accuracy by Lure Type and Count}

Table~\ref{tab:app_acc_by_lure} breaks base CSB accuracy down by lure type
and per-type lure count. The global readouts are systematically
lure-dependent: accuracy rises with additional target-shape and color
lures but drops as counterpart-shape lures are added, the signature of a
decision driven by scene-level color and shape statistics rather than by
the target conjunction. The foveated and oracle readouts are essentially
flat across all nine cells, staying above $97\%$ for every encoder and
probe.

\subsection{Training Protocols}

Encoder features are precomputed once and cached, so the encoder is
never part of the training graph. Only the readout (where applicable)
and the probe are trained, with five seeds (0--4) per configuration
and no data augmentation beyond each checkpoint's default
preprocessing.

\paragraph{Synthetic binding tasks (CSB and CSD).}
Both synthetic tasks use an identical recipe across all three
encoders and differ only in the number of classes ($C{=}6$ vs.\
$C{=}2$). The MLP probe is
$\mathrm{Linear}(D\!\to\!128)$--GELU--$\mathrm{Linear}(128\!\to\!C)$,
and the linear probe is a single $\mathrm{Linear}(D\!\to\!C)$, where
$D$ is the encoder's feature dimension ($1152$ for SigLIP and
SigLIP~2, $1024$ for CLIP). The foveated readout trains a single
query vector $q\in\mathbb{R}^{D}$ (initialized
$\mathcal{N}(0,0.02^2)$) with the patch tokens acting as their own
keys and values and no temperature scaling. Training uses the Adam
optimizer \cite{kingma2015adam} with default hyperparameters and no
weight decay, learning rate $10^{-3}$, batch size 512, 30 epochs, a
constant schedule, and unweighted softmax cross-entropy on the
$6{,}000/2{,}000/2{,}000$ train/val/test splits. The checkpoint with
the best validation top-1 accuracy is restored for evaluation.

\paragraph{GQA$^{*}$ color binding.}
The MLP probe is
$\mathrm{Linear}(d\!\to\!512)$--ReLU--Dropout--$\mathrm{Linear}(512\!\to\!10)$,
with the dropout rate selected per seed on validation from
$\{0.1,0.2,0.3,0.5\}$. The linear probe is a single layer without
dropout. The foveated readout trains a query map
$W_q\in\mathbb{R}^{d_{\mathrm{text}}\times d_{\mathrm{patch}}}$,
layer normalizations of the text embedding and of the patch tokens (a
per-token standardization of the features, applied only when
computing the attention scores, while pooling uses the unnormalized
patches), and a scalar temperature parameterized as $\log T$,
initialized at $\sqrt{d_{\mathrm{patch}}}$.
The text embedding $t(Q)$ comes from the text tower of the same
frozen checkpoint as the vision features (for CLIP,
$d_{\mathrm{text}}{=}768\neq d_{\mathrm{patch}}{=}1024$). Training
uses Adam with weight decay $10^{-4}$, learning rate $10^{-3}$
(identical across encoders), batch size 256, 50 epochs, constant
schedule, and cross-entropy with inverse-frequency class weights over
the 10 color classes ($\approx$6:1 imbalanced). Checkpoint of best validation paired accuracy used for testing.

\subsection{Dataset Construction Details}

Here, we explain how the three evaluation datasets are built.
For CSB we specify the shape vocabulary, the color palette, the
object placement process, and the generation of counterfactual
pairs and nuisance groups. For CSD we describe the size sampling and
placement changes of the color-free variant. For GQA$^{*}$ we
describe the filtering of scene-graph objects, the mining of
same-category color pairs, and the split protocol.

\paragraph{Color--shape binding (CSB).}
The 18 distractor-only shapes are the inverted, left-, and
right-rotated triangles, square, 4-, 5-, 6-, and 8-pointed stars,
4 semicircle orientations, plus, latin cross, chevron,
parallelogram, diamond, and ellipse. The palette is red
(\texttt{\#E51919}), orange (\texttt{\#FF7F00}), yellow
(\texttt{\#E5CC00}), green (\texttt{\#00B200}), blue
(\texttt{\#1919E5}), and purple (\texttt{\#990099}) on a white
canvas. All objects share one fixed size (half-extent $0.037$ of the
image width, $\approx$14\,px) and fixed canonical orientations, and
no rotation is sampled. Object centers are drawn uniformly with an
edge margin, placed by rejection sampling under a minimum
center-to-center distance of twice the shape extent (200 attempts per
object). An unplaceable distractor is dropped (metadata records
realized counts), and the scene is regenerated if the target fails to place. Lure counts are sampled i.i.d.\ per type,
$n_i\sim\mathrm{Uniform}\{1,2,3\}$. For each base scene, up to 5
counterfactuals (one per other class) are built by
histogram-preserving attribute swaps with a matched partner object: a
noise object carrying the needed color for color swaps, or a
counterpart-shape lure for shape swaps. A candidate is accepted if
all required partners exist, yielding 5, 3, or 1
counterfactuals per base scene and $9{,}366$ over the test split. Each of the 16 nuisance
variants fixes the target, $K$, and $\lambda$, and resamples all
non-target positions, shapes, and colors under the same role
constraints.

\paragraph{Crowded shape detection (CSD).}
CSD inherits the canvas, distractor vocabulary, fixed orientations,
and split sizes from CSB, including a 500-image clean $K{=}0$ split.
The target half-extent is $0.02$ of the image width ($\approx$8\,px),
distractor sizes are $1$--$11.5\times$ the target's, and all objects
are solid black. Placement is analogous to CSB but spaces per-shape
safety circles with a fixed 14\,px margin. An unplaceable distractor
is shrunk stepwise (never below $1\times$) rather than dropped, so
every scene retains its sampled $K$. There are exactly $2{,}000$
counterfactuals, one per test image, flipping the target shape in
place while preserving everything else.

\paragraph{GQA-derived color binding (GQA$^{*}$).}
We pool the images of the official GQA train and val scene-graph
releases. All questions are generated from the scene graphs (GQA's
own questions are not used). An object is color-bearing if both
bounding-box sides are $\geq 8$\,px and exactly one color from a
15-word vocabulary appears among its attributes (the match is
literal, with no synonym normalization, and multi-color annotations
are rejected). A category in an image yields a pair if it has exactly
two color-bearing instances with different colors whose box centers
are $\geq 20$\,px apart on the disambiguating axis (the axis of
larger center displacement). That axis determines the relation
(\texttt{left}/\texttt{right}/\texttt{above}/\texttt{below}).
Contributing images are split $80/10/10$ image-disjointly with both
members of a pair sharing a split ($17{,}994/2{,}266/2{,}242$
examples). The reported set keeps pairs whose two labels are both
among the ten most frequent colors: white, black, blue, red, brown,
green, gray, yellow, orange, and pink. Images are anisotropically
resized to $384\times384$ with boxes scaled accordingly. The oracle
input is the target's ground-truth box cropped from the resized image
and stretched to $384\times384$.

\subsection{Computing Infrastructure}

All experiments were run on a single Linux server (Ubuntu 24.04 LTS)
with 8$\times$ NVIDIA A40 GPUs (48\,GB each), an AMD EPYC 7763 64-core
CPU, and 2\,TB of RAM. Each experiment uses a single GPU. All
experiments use Python 3.11.15, PyTorch 2.6.0 (CUDA 12.4, cuDNN 9.2),
and Hugging Face \texttt{transformers} 5.12.1. Frozen encoder
checkpoints and their preprocessing are obtained from the Hugging Face
Hub via \texttt{transformers}: SigLIP
(\texttt{google/siglip-so400m-patch14-384}), SigLIP~2
(\texttt{google/siglip2-so400m-patch14-384}), and CLIP
(\texttt{openai/clip-vit-large-patch14-336}).

\input{tables/app/tab_app_promised}

\input{tables/app/tab_app_acc_vs_nsr}

\input{tables/app/tab_app_acc_by_lure}

%% file: 2-related.tex
\subsection{Related Work}

This subsection expands the discussion of related work deferred from
the main paper. We first review evidence that vision--language models
fail on compositional and spatially localized queries, and connect
these failures to classical accounts of the binding problem. We then
relate our diagnostic to work on dense patch-token representations,
attribute binding, question-conditioned localization in VQA, and
counterfactual evaluation protocols.

\paragraph{Spatial and compositional failures in VLMs.}
Large-scale vision--language models are often evaluated through image--text
matching, retrieval, or visual question answering, where a global visual
representation is compared with or fused into language
\cite{radford2021learning,zhai2023sigmoid}. A substantial body of work shows
that these models can recognize objects and attributes while failing on
compositional queries that require binding them to the correct referents.
Winoground tests whether models distinguish captions that swap object--relation
structure \cite{thrush2022winoground}; VL-Checklist and related benchmarks
probe object, attribute, and relation sensitivity \cite{zhao2022vl};
and bag-of-words analyses show that contrastive models often behave as if
captions and images were unordered collections of concepts
\cite{yuksekgonul2022and}. Spatial reasoning benchmarks reach similar
conclusions: models that perform well on broad image--text tasks often struggle
with left/right, above/below, containment, and other localized relations
\cite{liu2023visual,kamath2023whatsup,subramanian2022reclip}.

More recent work sharpens this diagnosis. Lewis et al. directly ask whether
CLIP binds concepts, finding that performance drops when multiple objects or
relations require structure-sensitive binding \cite{lewis2024does}. Tong et al.
construct CLIP-blind pairs and the MMVP benchmark, showing that visually
distinct images can be close in CLIP space and that these failures transfer to
multimodal LLMs \cite{tong2024eyes}. Rahmanzadehgervi et al. similarly argue
that modern VLMs remain brittle on simple visual tasks despite strong benchmark
performance \cite{rahmanzadehgervi2024vision}. Campbell et al. connect such
failures to the classical binding problem, arguing that multi-object reasoning
can induce representational interference when distinct entities share common
representational resources \cite{campbell2024understanding}. In 2025, new spatial
benchmarks such as SRBench continue to find that current VLMs perform near
chance on carefully isolated spatial-reasoning tasks
\cite{stogiannidis2025mind}. Our work agrees that localized binding is a
persistent failure mode, but asks a different mechanistic question: whether the
failure reflects missing information in the frozen encoder, or information lost
when patch tokens are collapsed into a global interface.

\paragraph{The binding problem and selective visual attention.}
The motivation for our synthetic task follows classical accounts of visual
binding. Feature-integration theory argues that conjunctions of separable
features, such as color and shape, require attention to bind features to the
same object \cite{treisman1980feature,treisman1982illusory}. Object-file theory
likewise proposes temporary, object-specific episodic representations in which
successive states of a perceived object are linked and integrated
\cite{kahneman1992reviewing}. Recent VLM work has revisited the binding problem
through the lens of serial and spatially organized processing. Campbell et al.\
interpret failures on multi-object tasks as consequences of the binding problem
and compare these failure modes to limitations of rapid feedforward processing
in human vision \cite{campbell2024understanding}. Izadi et al.\ propose VISER,
which combines simple visual scaffolds, such as horizontal or grid lines, with
prompts that encourage sequential, spatially grounded parsing
\cite{izadi2026visual}. Their results show that such input structuring can
mitigate binding-related failures across several visual-reasoning tasks.

Our foveated readout is related in spirit but differs in purpose. We do not
modify the input image, prompt a generative model, or train a new vision tower.
Instead, we keep the encoder fixed and train a single-query attention readout
over final patch tokens. The readout is therefore a diagnostic intervention: if
a small selector recovers the target conjunction while GAP and summary readouts
fail, then the relevant binding signal was present in the frozen spatial tokens
but suppressed by the global embedding interface.

\paragraph{Global embeddings v.s. dense patch-token representations.}
Contrastive image--text encoders are commonly optimized for global alignment:
an image is compressed into a class token, pooled output, or global average
before being compared with a text embedding. This makes the standard interface
well suited to image-level recognition and retrieval, but potentially lossy for
tasks whose label depends on a small region or on selecting one object among
many. Dense prediction work has long exploited the fact that CLIP-like models
retain useful patch-level information despite their image-level training.
MaskCLIP extracts dense labels from CLIP features without fully supervised
segmentation training \cite{zhou2022extract}; DenseCLIP and related methods
adapt CLIP to pixel- or region-level prediction \cite{rao2022denseclip};
and SCLIP modifies self-attention to improve dense vision--language inference
\cite{wang2024sclip}. Qiu et al. argue that CLIP contains useful
spatial correlations and propose distillation methods to preserve and refine
spatial awareness for open-vocabulary dense prediction \cite{qiu2025refining}.

This line of work is closely related to our claim that patch tokens contain
information not exposed by global embeddings. However, dense-prediction methods
typically aim to improve segmentation, detection, or open-vocabulary grounding,
often by changing attention, fine-tuning, distilling, or adding region-level
training objectives. Our goal is instead to isolate the evaluation interface.
We compare GAP, pretrained summary tokens, learned foveation, and oracle
target-token pooling on the same frozen encoder and the same downstream task.
The oracle readout estimates what is available in the frozen patch tokens when
location is supplied, while the learned fovea tests whether a lightweight
selector can recover that signal without masks at training time.

\paragraph{Attribute binding and the role of the interface.}
Recent compositionality work increasingly suggests that apparent global
failures may coexist with more structured information in parts of the model.
Koishigarina et al. find that CLIP behaves like a bag-of-words model
cross-modally but not necessarily unimodally: attribute--object binding
information can be present within individual image and text modalities, while
cosine alignment between global embeddings fails to use it correctly
\cite{koishigarina2025clip}.Uselis et al. similarly distinguish concept decodability from compositional
organization: individual concepts can be recoverable from learned features even
when models fail on unseen combinations, while stronger generalization emerges
alongside a more linearly factored feature geometry
\cite{uselis2025does}. Kamath et al. show that improvements on
compositional hard negatives can be overstated if evaluations do not also test
hard positives, motivating more careful minimal-pair protocols
\cite{kamath2024hard}. These findings are complementary to ours:
rather than asking only whether a model has a compositional representation, we
ask which readout exposes the relevant compositional signal.

\paragraph{Question-conditioned localization in VQA.}
Spatial attention has been central to image captioning and VQA since early
attention-based models, which use language or decoder state to weight image
regions before producing an answer or caption
\cite{xu2015show,anderson2018bottom}. Modern multimodal architectures likewise
provide language with access to multiple visual features, but through different
interfaces. Flamingo uses a Perceiver Resampler followed by gated
cross-attention in the language model \cite{alayrac2022flamingo}; BLIP-2 uses
a Q-Former whose learned queries cross-attend to frozen image features
\cite{li2023blip}; and LLaVA projects visual grid features into the language
embedding space and processes the resulting visual tokens together with text
tokens in the language model \cite{liu2023visualInstruction}. Our foveated
readout is not proposed as an alternative to these full multimodal
architectures. Rather, it is a diagnostic intervention that holds the vision
encoder fixed and isolates the value of question-conditioned visual selection
before spatial collapse. Our GQA-derived evaluation instantiates a minimal
localized binding problem in natural images: two paired questions share the
same pixels and object category but refer to different same-category instances
with different colors. In this setting, a question-independent global image
vector provides the same visual summary for both questions, whereas a
question-conditioned foveated readout can select different visual evidence
before pooling. The GQA experiment therefore serves as a natural-image
counterpart to the synthetic color--shape binding task.

\paragraph{Counterfactual diagnostics and nuisance variation.}
Several compositional benchmarks use hard negatives or minimal pairs to prevent
models from succeeding through single-concept recognition
\cite{thrush2022winoground,yuksekgonul2022and,kamath2024hard}. Our
synthetic setting extends this idea with matched label-changing
counterfactuals and label-preserving nuisance groups. The counterfactuals alter
the target conjunction while preserving nuisance statistics whenever possible;
the nuisance groups vary irrelevant scene content while keeping the label fixed.
This lets us measure not only accuracy, but also the nuisance-to-signal ratio
of a readout. The resulting analysis connects a qualitative failure---global
readouts ignore target edits under clutter---to a representation-level
mechanism: global pooling attenuates localized label-changing signal while
retaining variation from irrelevant objects. Thus, our contribution is not a
new benchmark alone, but a diagnostic framework for separating
representation failure from readout failure.

%% file: tables/app/tab_app_promised.tex
\begin{table*}[!h]
\centering
\small
\scalebox{1}{
\begin{tabular}{llccccc}
\toprule
& & \multicolumn{2}{c}{CSB} & \multicolumn{2}{c}{CSD}
& GQA$^{*}$ \\
\cmidrule(lr){3-4} \cmidrule(lr){5-6} \cmidrule(lr){7-7}
Config & Readout
& Clean & Paired-CF
& Clean & Paired-CF
& Paired \\
\midrule
SigLIP, Linear    & QO       & n/a & n/a & n/a & n/a & 2.8 $\pm$ 0.5 \\
SigLIP, Linear    & Summary  & 100.0 $\pm$ 0.0 & 2.8 $\pm$ 0.0 & 96.4 $\pm$ 1.4 & 47.2 $\pm$ 0.2 & 1.8 $\pm$ 0.3 \\
SigLIP, Linear    & GAP      & 99.8 $\pm$ 0.5 & 2.8 $\pm$ 0.1 & 86.8 $\pm$ 7.3 & 52.3 $\pm$ 0.3 & 1.8 $\pm$ 0.5 \\
SigLIP, Linear    & Foveated & 96.6 $\pm$ 3.1 & 94.1 $\pm$ 1.5 & 97.1 $\pm$ 0.3 & 94.2 $\pm$ 0.4 & 16.2 $\pm$ 0.4 \\
SigLIP, Linear    & Oracle   & 100.0 $\pm$ 0.0 & 99.3 $\pm$ 0.1 & 98.4 $\pm$ 0.2 & 97.5 $\pm$ 0.1 & 31.5 $\pm$ 0.9 \\
\midrule
SigLIP, MLP       & QO       & n/a & n/a & n/a & n/a & 2.3 $\pm$ 0.4 \\
SigLIP, MLP       & Summary  & 100.0 $\pm$ 0.0 & 3.2 $\pm$ 0.1 & 92.9 $\pm$ 9.4 & 53.6 $\pm$ 0.4 & 3.8 $\pm$ 0.6 \\
SigLIP, MLP       & GAP      & 100.0 $\pm$ 0.0 & 3.5 $\pm$ 0.2 & 49.2 $\pm$ 0.2 & 58.9 $\pm$ 0.4 & 4.3 $\pm$ 0.3 \\
SigLIP, MLP       & Foveated & 99.6 $\pm$ 0.2 & 93.5 $\pm$ 0.6 & 97.7 $\pm$ 0.1   & 93.7 $\pm$ 1.0 & 17.0 $\pm$ 0.7 \\
SigLIP, MLP       & Oracle   & 100.0 $\pm$ 0.0 & 99.3 $\pm$ 0.1 & 97.9 $\pm$ 0.1 & 97.9 $\pm$ 0.1 & 33.0 $\pm$ 0.7 \\
\midrule
CLIP, Linear      & QO       & n/a & n/a & n/a & n/a & 2.1 $\pm$ 0.6 \\
CLIP, Linear      & Summary  & 99.4 $\pm$ 0.7 & 5.7 $\pm$ 0.2 & 49.0 $\pm$ 0.0 & 48.0 $\pm$ 0.6 & 0.9 $\pm$ 0.3 \\
CLIP, Linear      & GAP      & 52.8 $\pm$ 2.0 & 2.2 $\pm$ 0.2 & 51.6 $\pm$ 2.0 & 31.1 $\pm$ 0.6 & 1.1 $\pm$ 0.2 \\
CLIP, Linear      & Foveated & 96.0 $\pm$ 2.0 & 99.1 $\pm$ 0.2 & 99.9 $\pm$ 0.2 & 97.8 $\pm$ 1.3 & 15.6 $\pm$ 0.8 \\
CLIP, Linear      & Oracle   & 100.0 $\pm$ 0.0 & 99.8 $\pm$ 0.1 & 100.0 $\pm$ 0.0 & 99.8 $\pm$ 0.1 & 29.4 $\pm$ 0.8 \\
\midrule
CLIP, MLP         & QO       & n/a & n/a & n/a & n/a & 2.4 $\pm$ 0.2 \\
CLIP, MLP         & Summary  & 96.9 $\pm$ 1.8 & 8.7 $\pm$ 0.3 & 49.0 $\pm$ 0.0 & 53.8 $\pm$ 0.5 & 4.2 $\pm$ 0.5 \\
CLIP, MLP         & GAP      & 45.9 $\pm$ 1.8 & 2.6 $\pm$ 0.2 & 49.0 $\pm$ 0.0 & 35.8 $\pm$ 0.2 & 3.3 $\pm$ 0.3 \\
CLIP, MLP         & Foveated & 97.2 $\pm$ 2.9 & 99.2 $\pm$ 0.1 & 98.1 $\pm$ 3.2 & 97.4 $\pm$ 0.9 & 17.7 $\pm$ 0.7 \\
CLIP, MLP         & Oracle   & 100.0 $\pm$ 0.0 & 99.8 $\pm$ 0.0 & 100.0 $\pm$ 0.0 & 99.8 $\pm$ 0.0 & 31.3 $\pm$ 0.9 \\
\midrule
SigLIP~2, Linear  & QO       & n/a & n/a & n/a & n/a & 2.3 $\pm$ 0.4 \\
SigLIP~2, Linear  & Summary  & 100.0 $\pm$ 0.0 & 2.6 $\pm$ 0.1 & 99.0 $\pm$ 0.4 & 51.4 $\pm$ 1.1 & 1.4 $\pm$ 0.3 \\
SigLIP~2, Linear  & GAP      & 99.2 $\pm$ 0.9 & 2.2 $\pm$ 0.1 & 96.8 $\pm$ 6.5 & 57.9 $\pm$ 0.1 & 1.0 $\pm$ 0.1 \\
SigLIP~2, Linear  & Foveated & 100.0 $\pm$ 0.0 & 97.5 $\pm$ 0.3 & 98.7 $\pm$ 0.1 & 97.3 $\pm$ 0.6 & 24.7 $\pm$ 1.1 \\
SigLIP~2, Linear  & Oracle   & 99.9 $\pm$ 0.1 & 99.9 $\pm$ 0.1 & 99.8 $\pm$ 0.2 & 98.9 $\pm$ 0.1 & 36.5 $\pm$ 0.3 \\
\midrule
SigLIP~2, MLP     & QO       & n/a & n/a & n/a & n/a & 2.3 $\pm$ 0.4 \\
SigLIP~2, MLP     & Summary  & 100.0 $\pm$ 0.0 & 3.2 $\pm$ 0.1 & 98.4 $\pm$ 2.2 & 64.3 $\pm$ 0.6 & 3.9 $\pm$ 0.4 \\
SigLIP~2, MLP     & GAP      & 99.6 $\pm$ 0.6 & 2.9 $\pm$ 0.1 & 98.8 $\pm$ 0.2 & 70.9 $\pm$ 0.5 & 3.9 $\pm$ 0.2 \\
SigLIP~2, MLP     & Foveated & 100.0 $\pm$ 0.0 & 97.5 $\pm$ 0.3 & 99.4 $\pm$ 0.5 & 97.8 $\pm$ 0.3 & 25.4 $\pm$ 1.2 \\
SigLIP~2, MLP     & Oracle   & 100.0 $\pm$ 0.1 & 99.9 $\pm$ 0.0 & 99.6 $\pm$ 0.2 & 99.0 $\pm$ 0.1 & 37.8 $\pm$ 1.0 \\
\midrule
Chance            &          & 16.7 & 2.8 & 50.0 & 25.0 & 1.0 \\
\bottomrule
\end{tabular}}
\caption{
Full readout comparison: clean and paired accuracy (\%) on the
color--shape binding task (CSB), the crowded shape-detection task
(CSD), and the GQA-derived localized color-binding task (GQA$^{*}$),
for the SigLIP, CLIP, and SigLIP~2 encoders with linear and MLP
probes. Clean accuracy is measured on the no-distractor split. Paired
accuracy is paired-counterfactual accuracy for CSB and CSD, and
paired accuracy for GQA$^{*}$. Each entry is the mean $\pm$ one
standard deviation over five training seeds. The SigLIP, MLP rows
correspond to the results reported in the main paper.
}
\label{tab:app_promised}
\end{table*}

%% file: tables/app/tab_app_acc_vs_nsr.tex
\begin{table*}[t]
\centering
\small
\begin{tabular}{llcccccccc}
\toprule
& & \multicolumn{2}{c}{$K$=10--19} & \multicolumn{2}{c}{$K$=20--29}
& \multicolumn{2}{c}{$K$=30--39} & \multicolumn{2}{c}{$K$=40+} \\
\cmidrule(lr){3-4} \cmidrule(lr){5-6} \cmidrule(lr){7-8}
\cmidrule(lr){9-10}
Config & Readout
& Acc & NSR & Acc & NSR & Acc & NSR & Acc & NSR \\
\midrule
SigLIP, Linear    & Summary
& 59.4 & 7.10 & 50.4 & 10.42 & 43.1 & 12.88 & 41.2 & 14.96 \\
SigLIP, Linear    & GAP
& 62.8 & 6.97 & 52.3 & 10.40 & 44.5 & 13.35 & 40.4 & 15.40 \\
SigLIP, Linear    & Foveated
& 99.0 & 0.31 & 98.7 & 0.30 & 98.2 & 0.29 & 97.1 & 0.30 \\
SigLIP, Linear    & Oracle
& 99.8 & 0.29 & 99.8 & 0.34 & 99.5 & 0.37 & 99.2 & 0.40 \\
\midrule
SigLIP, MLP       & Summary
& 61.9 & 7.10 & 56.4 & 10.42 & 47.7 & 12.88 & 42.2 & 14.96 \\
SigLIP, MLP       & GAP
& 65.7 & 6.97 & 58.4 & 10.40 & 50.4 & 13.35 & 44.8 & 15.40 \\
SigLIP, MLP       & Foveated
& 99.1 & 0.32 & 98.7 & 0.30 & 97.9 & 0.29 & 96.0 & 0.30 \\
SigLIP, MLP       & Oracle
& 99.8 & 0.29 & 99.8 & 0.34 & 99.6 & 0.37 & 99.1 & 0.40 \\
\midrule
CLIP, Linear      & Summary
& 68.9 & 6.27 & 58.0 & 9.38 & 51.2 & 12.33 & 48.3 & 13.29 \\
CLIP, Linear      & GAP
& 50.7 & 4.90 & 42.9 & 6.80 & 34.3 & 8.02 & 33.9 & 8.68 \\
CLIP, Linear      & Foveated
& 99.8 & 0.25 & 100.0 & 0.24 & 99.4 & 0.23 & 99.3 & 0.23 \\
CLIP, Linear      & Oracle
& 100.0 & 0.31 & 100.0 & 0.34 & 100.0 & 0.36 & 99.6 & 0.39 \\
\midrule
CLIP, MLP         & Summary
& 72.8 & 6.27 & 64.3 & 9.38 & 53.7 & 12.33 & 52.2 & 13.29 \\
CLIP, MLP         & GAP
& 57.6 & 4.90 & 50.8 & 6.80 & 41.9 & 8.02 & 37.7 & 8.68 \\
CLIP, MLP         & Foveated
& 100.0 & 0.27 & 99.9 & 0.26 & 99.8 & 0.25 & 99.3 & 0.25 \\
CLIP, MLP         & Oracle
& 100.0 & 0.31 & 100.0 & 0.34 & 100.0 & 0.36 & 99.6 & 0.39 \\
\midrule
SigLIP~2, Linear  & Summary
& 56.4 & 7.54 & 46.2 & 10.50 & 41.4 & 13.15 & 41.0 & 15.65 \\
SigLIP~2, Linear  & GAP
& 58.7 & 7.75 & 50.1 & 11.19 & 44.7 & 14.15 & 43.2 & 16.86 \\
SigLIP~2, Linear  & Foveated
& 100.0 & 0.17 & 99.6 & 0.16 & 98.8 & 0.17 & 98.4 & 0.18 \\
SigLIP~2, Linear  & Oracle
& 100.0 & 0.16 & 100.0 & 0.18 & 100.0 & 0.21 & 99.8 & 0.23 \\
\midrule
SigLIP~2, MLP     & Summary
& 63.3 & 7.54 & 52.8 & 10.50 & 48.5 & 13.15 & 43.9 & 15.65 \\
SigLIP~2, MLP     & GAP
& 65.3 & 7.75 & 56.1 & 11.19 & 49.3 & 14.15 & 46.4 & 16.86 \\
SigLIP~2, MLP     & Foveated
& 99.8 & 0.17 & 99.6 & 0.16 & 99.0 & 0.16 & 98.5 & 0.18 \\
SigLIP~2, MLP     & Oracle
& 100.0 & 0.16 & 100.0 & 0.18 & 100.0 & 0.21 & 99.9 & 0.23 \\
\bottomrule
\end{tabular}
\caption{
Base test accuracy (\%) and counterfactual nuisance-to-signal ratio
(NSR) on the color--shape binding task (CSB) by distractor-count bin
$K$, for all encoders and probes. Each entry is a mean over five training seeds. The
SigLIP, MLP values are those shown in the accuracy-versus-NSR figure
of the main paper.
}
\label{tab:app_acc_vs_nsr}
\end{table*}

%% file: tables/app/tab_app_acc_by_lure.tex
\begin{table*}[t]
\centering
\small
\begin{tabular}{llccccccccc}
\toprule
& & \multicolumn{3}{c}{TS} & \multicolumn{3}{c}{CS}
& \multicolumn{3}{c}{CL} \\
\cmidrule(lr){3-5} \cmidrule(lr){6-8} \cmidrule(lr){9-11}
Config & Readout
& 1 & 2 & 3 & 1 & 2 & 3 & 1 & 2 & 3 \\
\midrule
  SigLIP, Linear    & Summary
  & 36.0 & 49.7 & 59.4 & 59.5 & 47.6 & 38.7 & 39.4 & 47.7 & 58.5 \\
  SigLIP, Linear    & GAP
  & 34.6 & 53.1 & 61.8 & 60.9 & 49.7 & 39.6 & 43.7 & 49.9 & 56.3 \\
  SigLIP, Linear    & Foveated
  & 97.8 & 98.6 & 98.4 & 98.4 & 98.8 & 97.5 & 97.8 & 98.6 & 98.3 \\
  SigLIP, Linear    & Oracle
  & 99.3 & 99.6 & 99.8 & 99.4 & 99.5 & 99.7 & 99.6 & 99.5 & 99.5 \\
  \midrule 
  SigLIP, MLP       & Summary
  & 38.1 & 53.8 & 63.6 & 63.1 & 52.5 & 40.7 & 43.5 & 49.7 & 62.8 \\
  SigLIP, MLP       & GAP
  & 38.2 & 59.2 & 66.6 & 67.0 & 55.6 & 42.1 & 47.7 & 53.6 & 63.0 \\
  SigLIP, MLP       & Foveated
  & 97.3 & 98.2 & 98.1 & 98.2 & 98.3 & 97.2 & 97.3 & 98.6 & 97.7 \\
  SigLIP, MLP       & Oracle
  & 99.3 & 99.6 & 99.9 & 99.5 & 99.5 & 99.7 & 99.6 & 99.5 & 99.6 \\
  \midrule 
  CLIP, Linear      & Summary
  & 43.3 & 60.7 & 65.4 & 66.8 & 56.9 & 46.3 & 49.4 & 56.8 & 63.5 \\
  CLIP, Linear      & GAP
  & 33.6 & 41.8 & 45.7 & 48.6 & 39.1 & 33.8 & 33.5 & 40.4 & 47.5 \\
  CLIP, Linear      & Foveated
  & 99.5 & 99.8 & 99.6 & 99.7 & 99.5 & 99.7 & 99.3 & 99.7 & 100.0 \\
  CLIP, Linear      & Oracle
  & 99.7 & 100.0 & 100.0 & 100.0 & 99.9 & 99.9 & 99.9 & 99.8 & 100.0 \\
  \midrule                                                                                                                                                     
  CLIP, MLP         & Summary                                                                                                                                  
  & 46.8 & 65.1 & 70.0 & 70.2 & 62.5 & 49.7 & 53.7 & 61.0 & 67.5 \\
  CLIP, MLP         & GAP
  & 35.2 & 48.8 & 56.6 & 57.4 & 47.0 & 36.7 & 40.7 & 46.1 & 54.1 \\
  CLIP, MLP         & Foveated
  & 99.6 & 99.9 & 99.7 & 99.7 & 99.7 & 99.7 & 99.4 & 99.8 & 99.9 \\
  CLIP, MLP         & Oracle
  & 99.7 & 100.0 & 100.0 & 100.0 & 99.9 & 99.9 & 99.8 & 99.8 & 100.0 \\
  \midrule 
  SigLIP~2, Linear  & Summary
  & 35.6 & 46.8 & 56.1 & 56.5 & 46.7 & 35.9 & 39.4 & 46.7 & 52.7 \\
  SigLIP~2, Linear  & GAP
  & 37.6 & 49.7 & 59.9 & 58.7 & 50.5 & 38.7 & 42.0 & 48.9 & 56.8 \\
  SigLIP~2, Linear  & Foveated
  & 99.0 & 99.2 & 99.2 & 99.2 & 99.5 & 98.8 & 98.9 & 99.4 & 99.2 \\
  SigLIP~2, Linear  & Oracle
  & 100.0 & 100.0 & 99.9 & 100.0 & 100.0 & 99.9 & 99.9 & 100.0 & 100.0 \\
  \midrule 
  SigLIP~2, MLP     & Summary
  & 37.6 & 54.2 & 64.0 & 64.0 & 54.8 & 37.8 & 46.3 & 50.9 & 59.0 \\
  SigLIP~2, MLP     & GAP
  & 38.9 & 57.2 & 66.3 & 67.4 & 56.9 & 38.9 & 46.5 & 53.7 & 62.6 \\
  SigLIP~2, MLP     & Foveated
  & 99.0 & 99.5 & 99.2 & 99.5 & 99.6 & 98.6 & 99.1 & 99.2 & 99.3 \\
  SigLIP~2, MLP     & Oracle
  & 100.0 & 100.0 & 99.9 & 100.0 & 100.0 & 99.9 & 99.9 & 100.0 & 100.0 \\

\bottomrule
\end{tabular}
\caption{
Base test accuracy (\%) on the color--shape binding task (CSB) under
adversarial lures, by lure type---target-shape (TS),
counterpart-shape (CS), and color (CL)---and per-type lure count
(1--3), for all encoders and probes. Each entry is a mean over five
training seeds. The SigLIP, MLP values are those shown in the
accuracy-by-lure figure of the main paper.
}
\label{tab:app_acc_by_lure}
\end{table*}